\documentclass{article}
\pdfoutput=1
\usepackage{arxiv}
\usepackage[utf8]{inputenc} 
\usepackage[T1]{fontenc}    
\usepackage{url}            
\usepackage{booktabs}       
\usepackage{amsfonts}       
\usepackage{nicefrac}       
\usepackage{microtype}      
\usepackage{lipsum}
\usepackage{graphicx}
\usepackage{subfigure}
\usepackage{natbib}
\usepackage{algorithm}
\usepackage{algpseudocode}
\usepackage{nameref}

\title{Constructing compact signatures for individual fingerprinting of brain connectomes}

\author{
  Vikram Ravindra\\
  Department of Computer Science\\
  Purdue University\\
  W Lafayette, IN 47906 \\
  \texttt{ravindrv@purdue.edu} \\
   \And
  Petros Drineas\\
  Department of Computer Science\\
  Purdue University\\  
  W Lafayette, IN 47906 \\\texttt{pdrineas@purdue.edu} \\
   \And
  Ananth Grama\\
  Department of Computer Science\\
  Purdue University\\
  W Lafayette, IN 47906 \\\texttt{ayg@cs.purdue.edu} \\
}

\begin{document}
\maketitle
\begin{abstract}
Recent neuroimaging studies have shown that functional connectomes are unique to individuals, i.e., two distinct fMRIs taken over different sessions of the same subject are more similar in terms of their connectomes than those from two different subjects. In this study, we present significant new results that identify, for the first time, specific parts of resting state and task-specific connectomes that code the unique signatures. We show that a very small part of the connectome codes the signatures. A network of these features is shown to achieve excellent training and test accuracy in matching imaging datasets. We show that these features are statistically significant, robust to perturbations, invariant across populations, and are localized to a small number of structural regions of the brain. Furthermore, we show that for task-specific connectomes, the regions identified by our method are consistent with their known functional characterization.  We present a new matrix sampling technique to derive computationally efficient and accurate methods for identifying the discriminating sub-connectome and support all of our claims using state-of-the-art statistical tests and computational techniques.
\end{abstract}


A number of functional MRI studies focus on differences between sets of cohorts (e.g., healthy vs. diseased). These studies characterize an invariant signal within each cohort and identify significant changes in these signals across cohorts. Invariant signals within cohorts are designed to suppress individual-level heterogeneity of the functional connectomes, and can be computed using methods that range from simple averaging to sophisticated matching techniques, to identify statistically significantly conserved components within the cohort.

Clinically observed functional and/or structural variability between individual brains typically manifests in structural and functional connectomes. It has been shown that functional connectomes of an individual express higher degree of similarity than connectomes of different individuals.
It has also been shown that restricting such analyses to regions constituting well-known networks such as fronto-parietal networks, amplifies the individual-specific signature. This suggests that the signal corresponding to identifiability is spatially selective in its expression.

In this paper, we present a novel technique for identifying {\em interpretable} individual-specific signatures in resting and functional brain networks. This is in contrast to prior efforts, which computed individual signatures as low-rank representations of the matrix corresponding to the connectome. While prior results demonstrated the existence of the signatures, the computed signatures are not physiologically relevant or even interpretable, since they do not map back to regions of the brain. In contrast, we identify a highly accurate and compact set of features from the connectome that map directly to regions of the brain. By not restricting our search for features to specific regions of the brain, we are able to identify a comprehensive, yet highly specific set of features. We show that these sets yield stronger discriminatory signals, resulting in higher accuracy of identification.  We also demonstrate the use of our method to isolate signatures that encode information specific to the task being performed by an individual. We demonstrate the statistical significance and robustness of our signatures, and show that the signatures correspond to regions of the brain that are consistent across subjects.

\section*{Results}
The goal of this effort is to characterize the uniqueness of the functional connectome of an individual in terms of its accuracy, underlying features, and associated task-specificity. We define the innate uniqueness property as identifiability. We define a \emph{signature} as a pattern that encodes identifiability. In this effort, we isolate small regions on the human cortex that strongly express these signatures. In this context, we use the terms \emph{signatures} and \emph{fingerprints} interchangeably. We present detailed experiments to characterize the accuracy of these signatures in terms of their ability to identify individuals, as well as the task they are performing.

We used the functional MRI images of the unrelated subset of subjects, acquired as part of the WU-Minn Project by the Human Connectome Project (HCP) Consortium. The images were acquired over two sessions on separate days. The first day included a resting session (REST1), followed by working memory (WM), motor task (MOTOR) and language (LANG). The second day included another resting session (REST2), social cognition (SOCIAL), relational processing (RELATION), and emotion processing (EMOTION). Each of the rest and task sessions included a left-right (LR) encoding and a right-left (RL) encoding. We follow the the minimum pre-processing pipeline designed by HCP. We used the Atlas designed by  \cite{Glasser16} to parcellate the cortex into 360 anatomically and neuro-physiologically relevant regions and constructed region-wise similarity matrices. We then vectorize these correlation matrices, after discarding symmetric components (see Online Methods for details). 

In our experiments, we use two groups ($G_1$ and $G_2$) of connectomes. Each group has exactly one data point per subject. Our task is to successfully match the connectivity matrices belonging to a subject across the two groups. In the first set of experiments, where we use resting state connectivity matrices, the two groups have correlation matrices from REST1 and REST2, respectively. In the experiment that investigates identifiability of individuals while performing tasks, the first group consists of LR encoding and the second group consists of RL encoding. 
In the final experiment that focuses on task identifiability, the first group consists of REST1 and the LR encodings of seven tasks, whereas the second group consists of REST2 and the RL encodings of the same seven tasks. Each group matrix is constructed by stacking the vectorized correlation matrices. Thus each data point in this matrix corresponds to a subject and each feature is a measure of coherence in activation between two regions. In all the experiments that follow, we use the principal features subspace method, which uses \emph{leverage scores} to select features (see Online Methods for more details).

\subsection*{A small set of features encode individual signatures}

Our first set of experiments is designed to demonstrate our key result, that a small number of features encode resting-state signatures that are largely unique to individuals. This is in contrast to prior techniques that do not identify either structural or functional features associated with individual signatures, just that such signatures exist in the aggregate datasets.

To demonstrate this result, we use the group matrix $G_1$ to identify features corresponding to the top 100 leverage scores (from among over 64K scores). We show later 
that accuracy of identification saturates at less than 100 features. Then, pairwise correlation scores are calculated between data points of the two groups in this restricted feature-space. The purpose of this experiment is to demonstrate that two Functional Connectomes of the same subject are more similar than two Functional Connectomes belonging to different subjects using {\em only } this small feature set.

We define accuracy of identification as the percentage of instances where individuals are correctly matched across the two groups, $G_1$ and $G_2$. Note that a match is identified as the highest correlation of an individual in $G_2$ over all individuals in $G_1$. To ensure robustness, we repeat this experiment for 1000 different (random) subsets of subjects. 
Table~\ref{leverage_split_tbl} presents accuracy for subsets of different sizes. The accuracy obtained in the reduced space is comparable to results obtained by using the full correlation matrices, or by using a low rank matrix approximation. 

To characterize the statistical significance of the selected features, we repeat the experiment with the same number of features chosen uniformly at random $10^6$ times. The average accuracy was much lower (approximately 55\%, see Table~\ref{feat_sel_tbl}). None of these instances yielded accuracy values higher that that for our leverage-score based feature selection. This suggests high statistical significance of our identified features (empirical $p$-value$< 10^{-6}$).   

\begin{table*}[ht]
\centering
\begin{tabular}{ |c|c|c| } 
\toprule
Training/test split & Training Set Accuracy(\%, mean $\pm$ std) & Test Set Accuracy(\%, mean $\pm$ std) \\
\midrule
 80/20 & $96.23 \pm 2.24$  &  $93.11 \pm 3.61$ \\
 50/50 & $96.30 \pm  2.59$ &  $92.94 \pm 3.82$ \\
 30/70 & $96.81 \pm 3.07$ & $ 90.23 \pm 4.30$ \\
 20/80 & $97.01 \pm 3.22$ & $87.60 \pm 5.27$ \\
 10/90 & $97.72 \pm 2.65$ & $81.86 \pm 7.15$ \\
 \bottomrule
  \end{tabular}
  \bigskip
  \caption{{\bf Accuracy of principal features subspace method for different splits of training and test samples.} The results show consistently high training accuracy across a range of training set sizes. As expected, training accuracy increases with smaller training sets; however, test accuracy goes down as training set size is reduced, due to overfitting.}
  \label{leverage_split_tbl}
\end{table*}

\subsection*{The feature set is robust across individuals}
\label{subsec:hyp_2}
Our first set of experiments identified a set of features from individuals in Group 1, and used these features to draw correspondence (identification) between the two groups. Note that in these experiments, the identification of signatures and identification of subjects is on the same set of individuals. In our second set of experiments, we show that this feature-set is invariant across individuals. Specifically, we identify features using one set of subjects (features with highest leverage scores in the training set), and use these features to draw correspondences between a distinct set of subjects (the test set) across the two groups. The result of this experiment is presented in Table \ref{feat_sel_tbl}. The high test accuracy indicates that features relevant for the training set are also discriminative for the test set (p-value: $< 10^{-6}$). This suggests an anatomical and/or physiological basis for the features selected.

\begin{table*}[ht]
\centering
\begin{tabular}{ |c|c|c| } 
\toprule
Feature Selection Method & Training Set Accuracy(\%, mean $\pm$ std) & Test Set Accuracy(\%, mean $\pm$ std) \\
\midrule
 Leverage Score & $96.23 \pm 2.24$  &  $93.11 \pm 3.61$ \\
 Random & $54.62 \pm 7.76$ & $54.52 \pm 7.52$ \\
 \bottomrule
 \end{tabular}
 \bigskip
\caption{{\bf Performance of leverage-score based feature selection technique, compared to selecting features uniformly at random }. The significantly higher test and training set accuracy demonstrates the performance of our principal subspace feature selection method.}
\label{feat_sel_tbl}
\end{table*}

To further highlight the implication of this result, it is possible to apply a computational procedure to extract features from training and test sets. The computational procedure may be a complex function defined over all features. This is in fact the case with prior methods. In contrast, what our experiment shows is that no such computational procedure is required -- rather, a robust selection of sample-invariant features is highly discriminative and accurate.

\subsection*{The size of the discriminating feature set is compact}
\label{subsec:hyp_3}
\begin{figure}
\begin{center}
\includegraphics[scale=0.5]{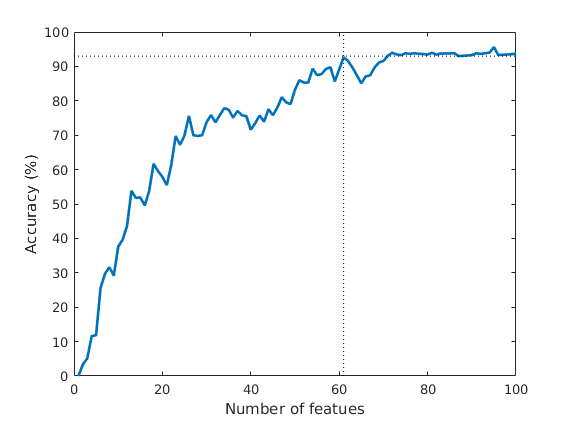}
\end{center}
\caption{Variation of test accuracy as a function of feature set size. The rapid convergence in test accuracy demonstrates that a small set of features codes the discriminating signature.}
\label{min_feats}
\end{figure}

Our first set of experiments demonstrated that a small set of 100 features is robust and statistically significant in its discriminative power. We now seek to determine the smallest such set that can code individual signatures with high accuracy. We design an experiment in which the number of features is successively increased (in order of decreasing leverage scores) and the test accuracy is measured at each step. The result of this experiment is presented in Figure~\ref{min_feats}. We note that the test accuracy essentially plateaus using 61 features at $93\%$. The maximum accuracy ($95.5\%$) is achieved using 95 features.

It is important to note that this experiment relies on a leverage score ordering of the features. It is indeed possible that an alternate ordering may yield a smaller feature set. However, our results on the suitability of leverage scores for feature selection, combined with the rapid plateauing of accuracy, strongly suggest that a feature set comprised of 61 to 95 features (from among over 64K features) is enough to code individual signatures with high accuracy.

\subsection*{A small number of structural regions encode the signature}

Our previous experiments used entries in the correlation matrix (edges in correlation network among regions of the brain) as features. We now investigate whether a small number of structural features (regions in the brain) are capable of coding individual signatures with high discriminating power.

\begin{figure*}
\centering
\includegraphics[scale=0.3]{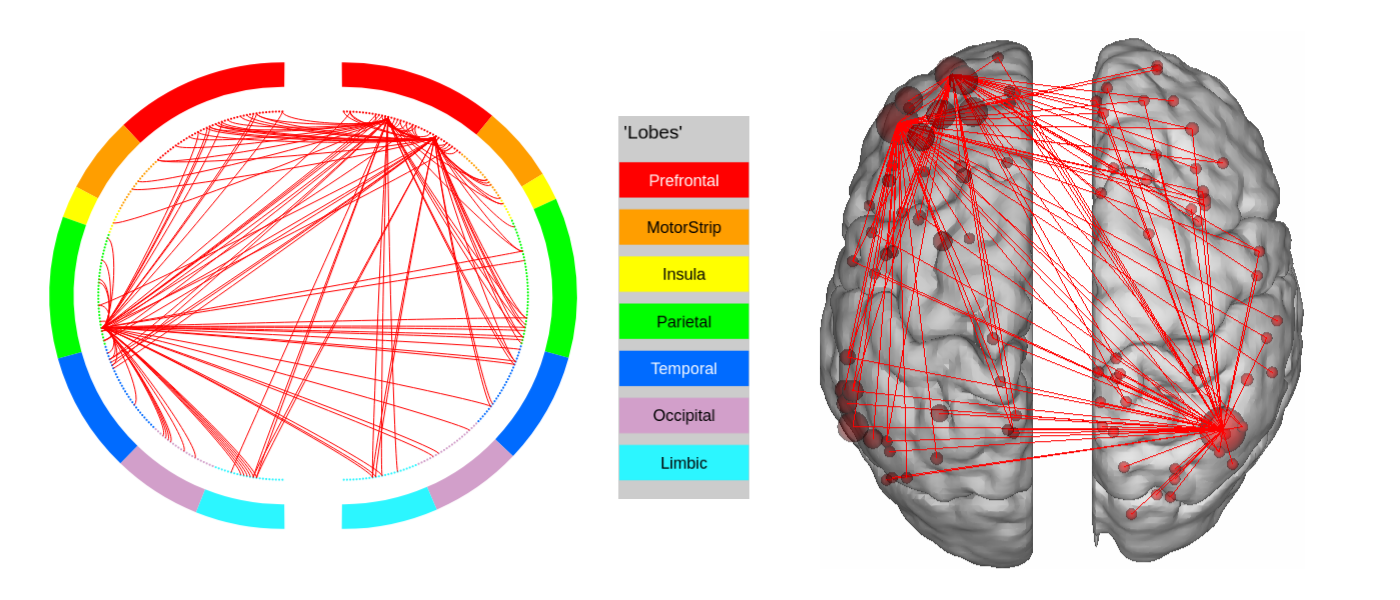}
\caption{{\bf High confidence edges that encode resting-state signature in the human connectome.} The connectivity map shows that the signature is strongly expressed in the prefrontal cortex and the parietal cortex. For illustrative purposes, we show edges when at least one terminal node has a degree of 30.}
\label{fig:resting_results}
\end{figure*}

In this experiment, we compute the top 100 features with the highest leverage scores for a randomly selected subset of 80 subjects. We repeat this procedure 1000 times. We then extract the high confidence features using a hyper-geometric p-value cutoff of $10^{-20}$. The results are shown in Fig~\ref{fig:resting_results}. We observe that the prefrontal cortex and the parietal regions strongly encode the signature. This is in agreement with \cite{Finn15}, who showed that the fronto-parietal network has the most discriminating power. However, in contrast to previous methods, we are able to make this observation solely by analyzing the resting state connectivity matrices of one session in the training set.

In order to find the regions based on our parcellation scheme, we select the regions that are over-represented in the top features across all 1000 tests (hypergeometric p-value $< 10^{-20}$). This process resulted in 12 high-confidence regions, which, as we show, encode the signature. To do this, we restricted ourselves to work with the $24 \times 24$ connectivity matrix, corresponding to both hemispheres of the previously identified regions. We then constructed the vectorized representation of this smaller matrix for the test subjects and found the accuracy to be $94.05$ ($\pm 1.22$). In comparison, identifiability accuracy for 12 randomly chosen regions was $41.47$ ($\pm 12.25$, for p-value $< 10^{-6}$). The regions in this set are listed in Table~\ref{top_regions_tbl}. Overall, the signature expressing regions occupy about 4.5\% of the cortex.

Structurally, five of the 12 regions (namely Area Anterior 9-46v, Area 9-46v, Inferior 6-8 transitional area, Area Anterior 10p, and Area 9 anterior) belong to the Dorsolateral Prefrontal Cortex. Furthermore, Intraparietal dorsal, Intraparietal 0, and Lateral Intra Parietal Ventral belong to the Parietal Cortex (see Supplementary Material in~\cite{Glasser16}). This shows that regions coding the signature are physiologically localized.

\begin{table*}[ht]
\centering
\begin{tabular}{|c|c|c|c|}
\toprule
Area Number & Area Notation & Region\\
\midrule
141 & Temporo-Parieto-Occipital Junction 3 & Parietal\\
11 & Premotor Eye Field & Premotor\\
95 & Intraparietal dorsal & Parietal\\
85 & Area Anterior 9-46v & Prefrontal\\
86 & Area 9-46v & Prefrontal\\
97 & Inferior 6-8 transitional area & Prefrontal\\
146 & Intra Parietal 0 & Parietal\\
48 & Lateral Intra Parietal Ventral & Parietal\\
89 & Area Anterior 10p & Prefrontal\\
87 & Area 9 anterior & Prefrontal\\
137 & PHT & Lateral Temporal\\
79 & IFJa & Inferior Frontal\\
\bottomrule
\end{tabular}
\bigskip
\caption{{\bf Top brain regions associated with discriminating features for resting-state connectomes.} We observe that most of the signature expressing regions are in the prefrontal and the parietal cortex.}
\label{top_regions_tbl}
\end{table*}

\subsection*{Our method significantly outperforms state of the art methods in terms of test accuracy}
\label{sec:results_comparison}

In this set of experiments, we compare the accuracy of our method with other state-of-the-art techniques. ~\cite{Finn15} compute all pairwise Pearson correlation coefficients, where  each data point is represented by all elements of the upper diagonal of the time-series correlation matrix. We randomly choose 50 out of 100 subjects and compute pairwise similarity; the accuracy of Finn et al.'s method is $88.65$ ($\pm 1.76$) averaged over 1000 trials.
\cite{Amico17} first denoise the data by retaining a subset of the principal components. As before, we randomly choose 50 subjects and compute the training set accuracy (see Table~\ref{svd_tbl}). We observe that removing the top few principal components helps to improve accuracy, since these components correspond to signals that are common to all subjects, as noted in~\cite{Amico17}. The test set accuracy is computed by removing the top principal components of the training dataset from the test dataset. This reveals that the common signal of the training dataset is an artifact of those particular subjects, and not a pattern that provides significant insights about the functional behaviour of the brain. In all cases, our method yields comparable training set accuracy, significantly higher test set accuracy, while relying on a small set of structurally and functionally interpretable features. 

\begin{table*}[ht]
\centering
\begin{tabular}{|c|c|c|}
 \toprule
 Principal Components & Training Set Accuracy(\%, mean $\pm$ std) & Test Set Accuracy(\%, mean $\pm$ std)  \\
 \midrule
 All (Same as \cite{Finn15}) & $88.65 \pm 1.76$ & $88.98 \pm 1.72$ \\ 
 2:end & $94.30 \pm 1.35$ & $71.76 \pm 8.76$ \\ 
 11:end & $96.74 \pm 1.00$ & $69.61 \pm 8.94$ \\
 21:end & $95.03 \pm 1.90$ & $69.44 \pm 8.99$ \\
 31:end & $71.97 \pm 6.08$ & $68.95 \pm 9.07$ \\
 41:end & $2.77 \pm  1.74$ & $65.70 \pm 9.59$ \\
 Principal Features Subspace & $96.23 \pm 2.24$  &  $93.11 \pm 3.61$ \\
 \bottomrule
 \end{tabular}
 \bigskip
 \caption{{\bf Comparison of training and test accuracy for various methods.} Numbers in the first row use the entire matrix; the next five rows use a subset of the principal components; the last row corresponds to our method. Our method achieves almost optimal training set accuracy and significantly better test set accuracy compared to competing methods.} 
\label{svd_tbl}
\end{table*}

\subsection*{Individuals performing tasks can be distinguished}

\begin{figure*}
    \centering
    \includegraphics[scale=0.5]{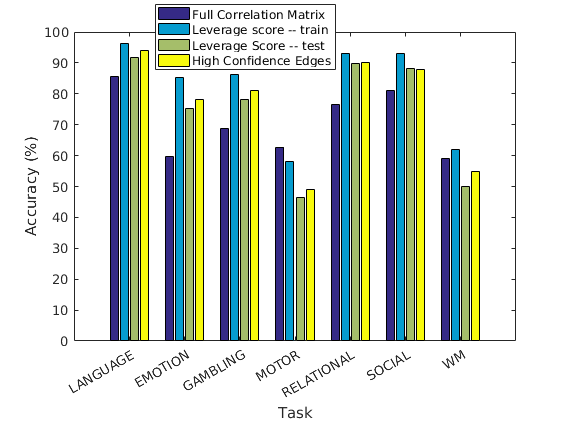}
    \caption{{\bf Prediction accuracy of individuals performing tasks.} }
    \label{fig:taskwise_accuracy}
\end{figure*}

In this experiment, we demonstrate that our method for finding discriminating edges extends to the case of task-based functional networks. For each task, we construct the two group matrices and partition subjects into training and test sets as described before. We compute the leverage scores separately for each task using the respective training subsets, restrict the feature space to top leverage scores, and report predictions on the test set in the same manner as in the resting state experiments. The results (averaged over 1000 trials) are shown in Figure~\ref{fig:taskwise_accuracy}. The performance of our approach is comparable to using the full correlation matrix in all but two tasks, namely MOTOR and WM. This indicates that our method recognizes features that important for most tasks. As before, we localize the features to physical regions on the brain. In order to do this, we only use features whose leverage scores are consistently sufficiently high to remain in the reduced feature space across all 1000 trials (hyper-geometric p-value $< 10^{-10}$).

\begin{figure*}
    \centering
    \includegraphics[scale=0.3]{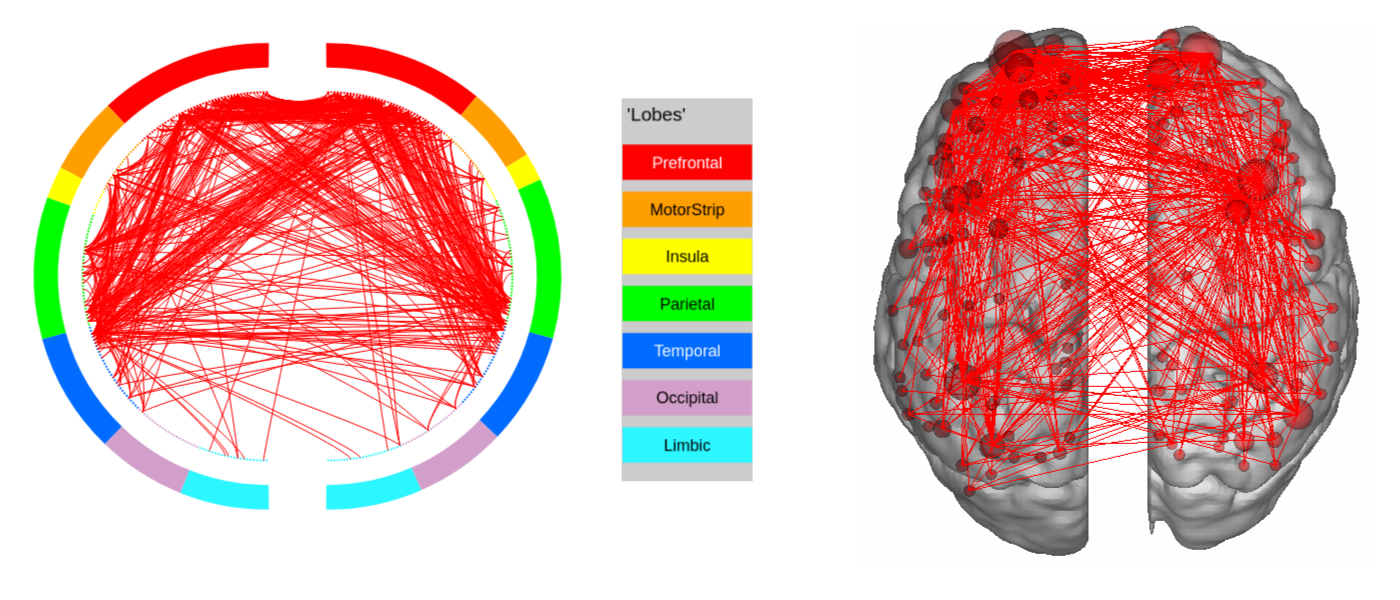}
    \caption{{\bf High confidence edges that encode signature while performing the language task of HCP.} The regions with high edge density are in good agreement with the regions of interest (ROIs) for language \cite{Barch13}.  For illustrative purposes, we show edges when at least one terminal node has a degree of 30.}
    \label{fig:LANGUAGE_regions}
\end{figure*}

We discuss the results for each task. In the LANGUAGE task, the statistically significant regions include the Inferior Frontal Cortex and surrounding regions (Orbital and Frontal Cortex and Dorsolateral Prefrontal Cortex). They also include regions in the Anterior Cingulate and Medial Prefrontal Cortex, as well as areas that surround the Superior Temporal Cortex (the Medial and Lateral Temporal Cortex). These results are consistent with the regions of interest for LANGUAGE, identified by \cite{Barch13}. Furthermore, the task involves auditory inputs, and requires communication via pressing a button. This explains the Premotor Cortex and the Auditory Association Cortex being recognized as regions of interest by our approach. The high confidence edges for the task are shown in Figure \ref{fig:LANGUAGE_regions}.

\begin{figure*}
    \centering
    \includegraphics[scale=0.3]{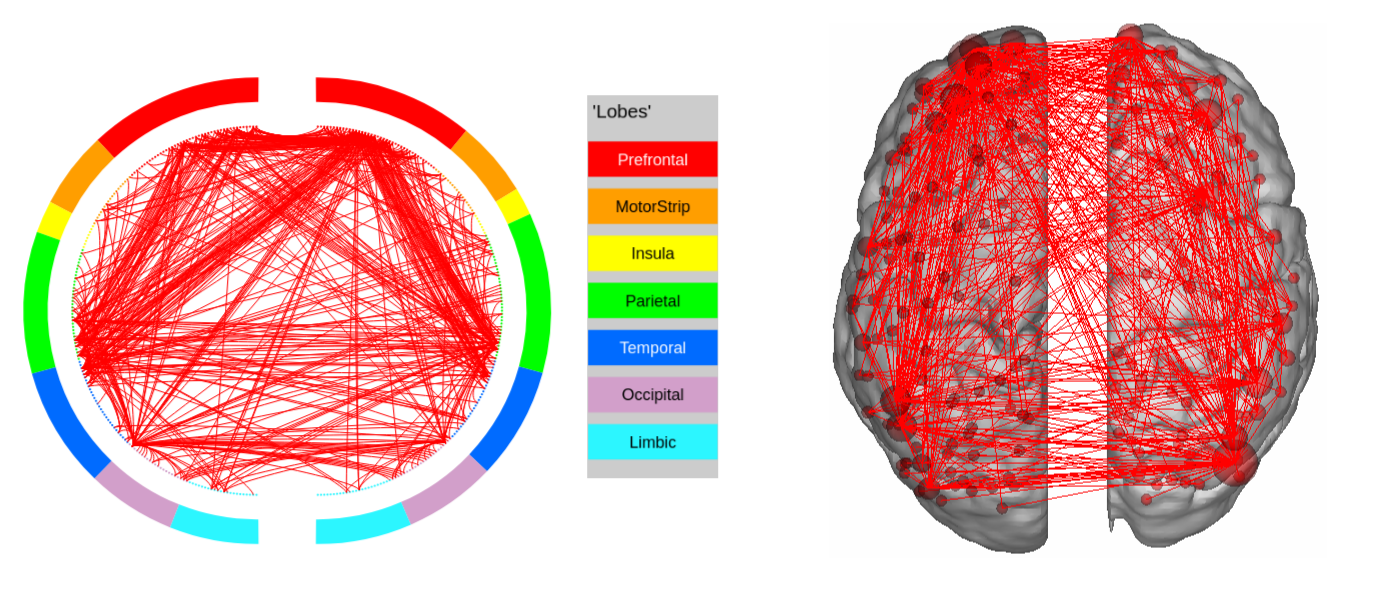}
    \caption{{\bf High confidence edges that encode signature while performing the emotion processing task of HCP.} The regions with high edge density are in good agreement with the ROIs for emotion processing \cite{Barch13}.  For illustrative purposes, we show edges when at least one terminal node has a degree of 30.}
    \label{fig:EMOTION_regions}
\end{figure*}

\begin{figure*}
    \centering
    \includegraphics[scale=0.3]{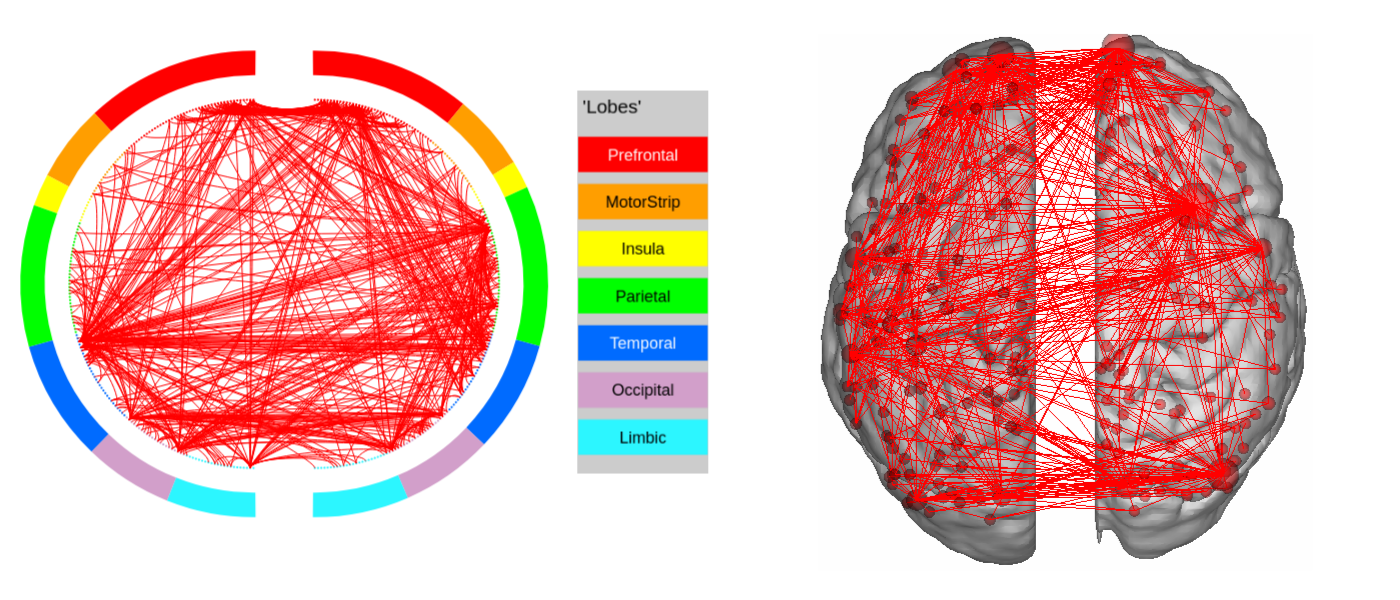}
    \caption{{\bf High confidence edges that encode signature while performing the gambling task of HCP.} The regions with high edge density are in good agreement with the ROIs for gambling \cite{Barch13}.  For illustrative purposes, we show edges when at least one terminal node has a degree of 30.}
    \label{fig:GAMBLING_regions}
\end{figure*}

\begin{figure*}
    \centering
    \includegraphics[scale=0.3]{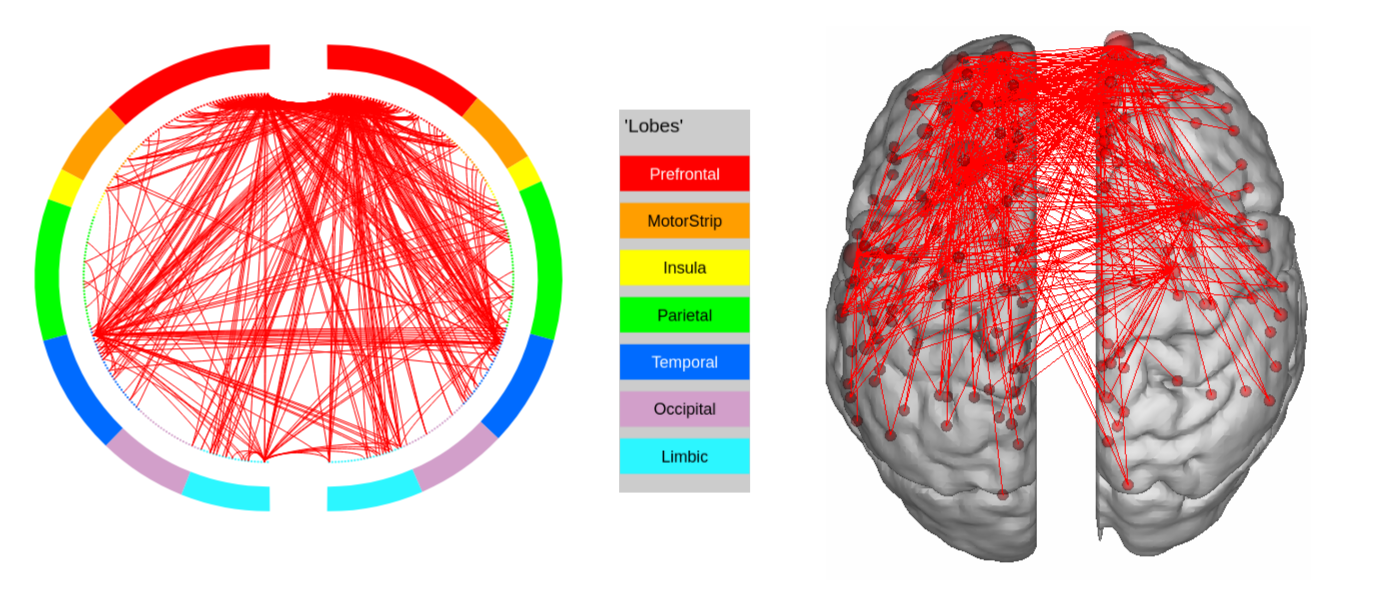}
    \caption{{\bf High confidence edges that encode signature while performing the language task of HCP.} Our method does not find the regions implicated in this case.  For illustrative purposes, we show edges when at least one terminal node has a degree of 30.}
    \label{fig:MOTOR_regions}
\end{figure*}

\begin{figure*}
    \centering
    \includegraphics[scale=0.3]{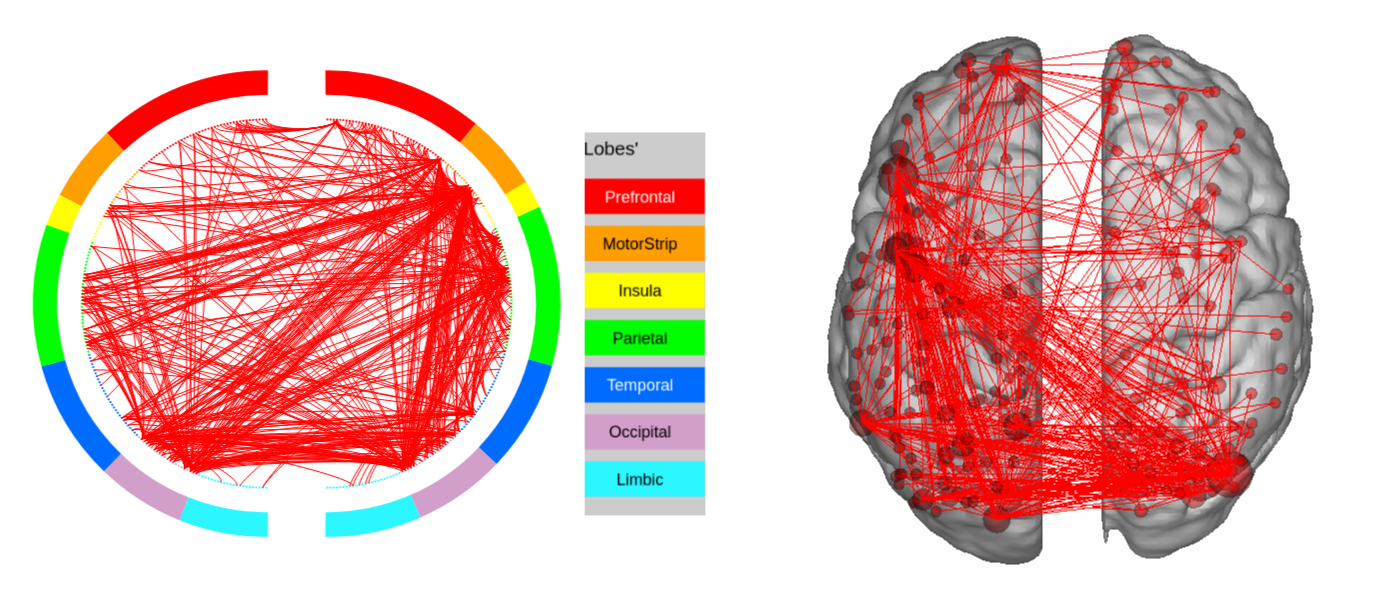}
    \caption{{\bf High confidence edges that encode signature while performing the relational processing task of HCP.} The regions with high edge density are in good agreement with the ROIs for relational processing.  For illustrative purposes, we show edges when at least one terminal node has a degree of 30.}
    \label{fig:RELATIONAL_regions}
\end{figure*}

\begin{figure*}
    \centering
    \includegraphics[scale=0.3]{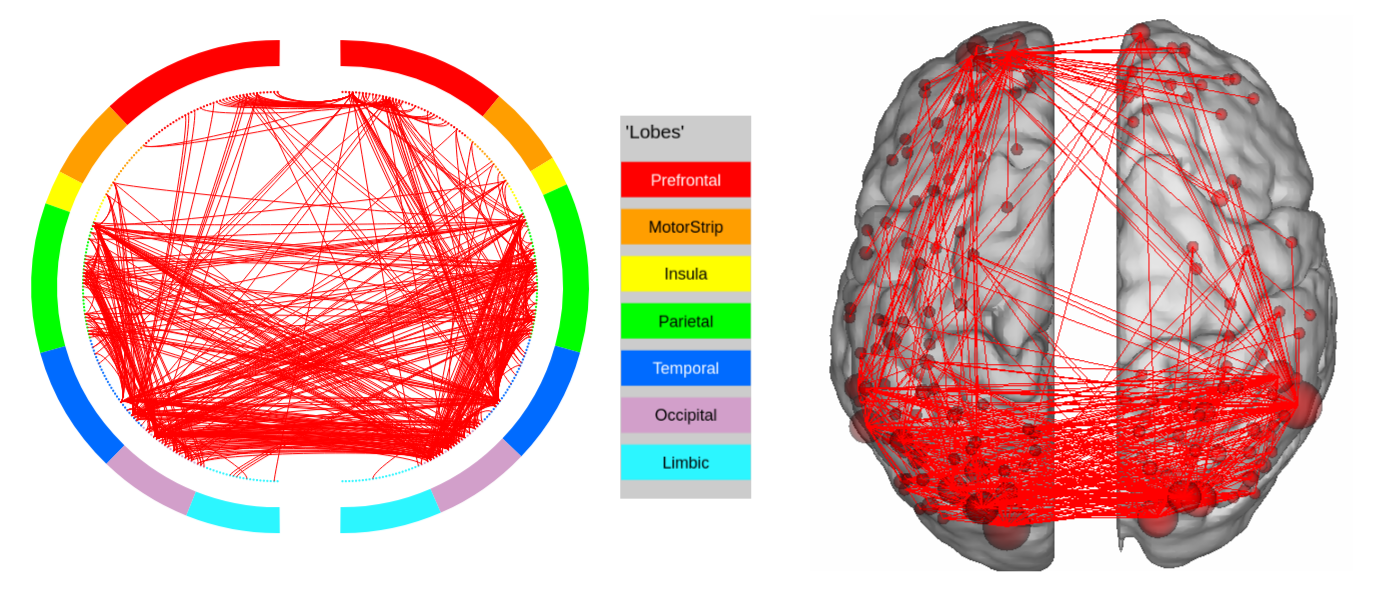}
    \caption{{\bf High confidence edges that encode signature while performing the social processing task of HCP.} The regions with high edge density are in good agreement with the ROIs for social processing.  For illustrative purposes, we show edges when at least one terminal node has a degree of 30.}
    \label{fig:SOCIAL_regions}
\end{figure*}

In EMOTION, the regions implicated in the signature are found in the Prefrontal Cortex and neighboring regions (Inferior Frontal Cortex, Orbital and Polar Frontal Cortex), as well as areas surrounding the Insula (Lateral and Temporal Parietal Cortex). These results are consistent with the results of ~\cite{Barch13}, except for the fact that we do not identify the Medial Temporal Cortex as a region of interest. The connectivity graph for EMOTION is shown in Figure \ref{fig:EMOTION_regions}. In GAMBLING, we find the expected regions of interest in the Prefrontal cortex, as well as the Orbito-frontal cortex. The other area of interest is the Sub-cortical Striatum, which is not included in our parcellation scheme. In this experiment, the subjects are shown visual stimuli and are required to communicate their decision by pressing a button (Figure~\ref{fig:GAMBLING_regions}). This explains the fact that the regions in Visual and Motor Cortex are being recognized as potential regions of interest. The results for SOCIAL tasks are shown in Figure \ref{fig:SOCIAL_regions}: we identify all relevant regions of interest (Temporal Parietal Junction and Medial Prefrontal Cortex, along with the visual cortex and the motor cortex). In the RELATIONAL task, we identify regions in the prefrontal cortex, but also in the parietal cortex, as shown in Figure \ref{fig:RELATIONAL_regions}. In the MOTOR task (Figure \ref{fig:MOTOR_regions}), we do not identify the regions of interest (Somatosensory and Motor cortex). This explains the poor prediction accuracy for this task. Similarly, we do not identify the Entorhinal Cortex or the Hippocampus in WM, which again explains the poorer prediction accuracy for this task.

While the regions obtained by our method broadly agree with the regions that are expected to be activated during respective tasks, we note that a general pipeline of tasks for fMRI would factor in additional information, such as presence/absence of various stimuli, duration of each block, and performance metrics associated with the tasks performed. The connectivity profiles of each of the tasks are sufficiently distinct, suggesting that we can identify tasks, which is indeed our main conclusion. 

\begin{figure*}
    \centering
    \includegraphics[scale=0.3]{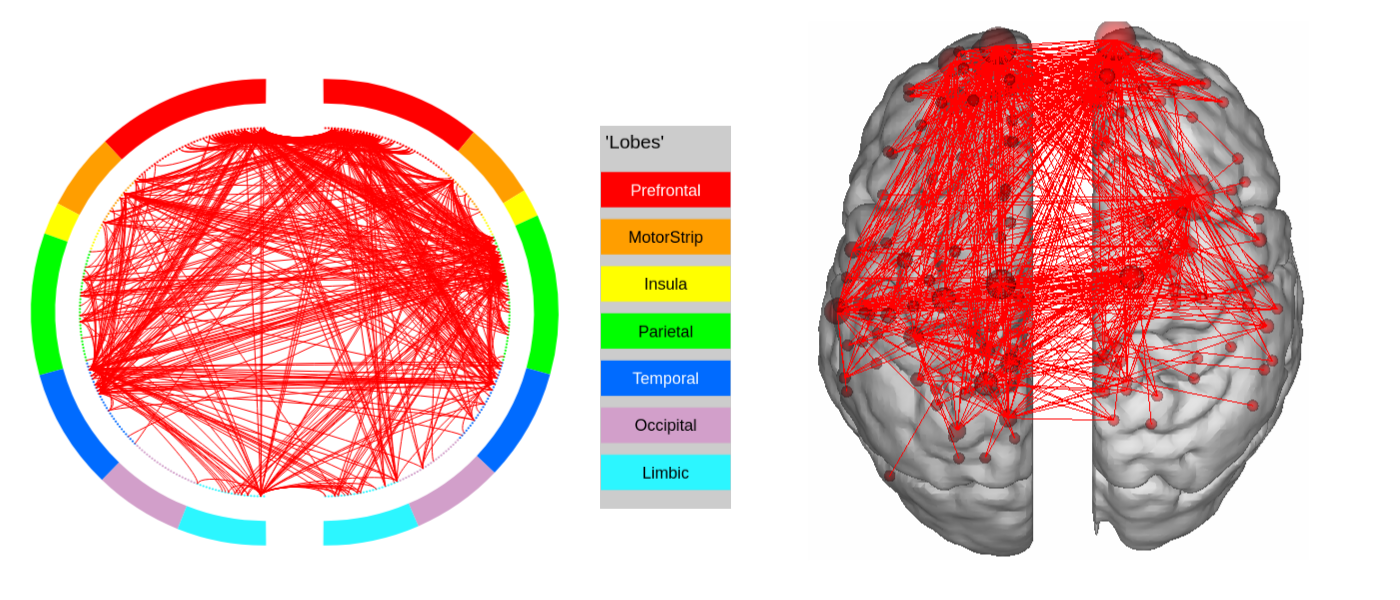}
    \caption{{\bf High confidence edges that encode signature while performing the working memory task of HCP.} Our method does not find the regions implicated in this case.  For illustrative purposes, we show edges when at least one terminal node has a degree of 30.}
    \label{fig:WM_regions}
\end{figure*}

\subsection*{Task-identifying signatures are strongly localized to the Occipital Lobe}

In this final experiment, we aim to identify the tasks performed by an individual. For each subject, we create two group matrices. The first group matrix contains the LR encoding of all tasks and the second group matrix contains the corresponding RL encoding. Here, a match is successful if the same tasks across the two groups are more similar to each other than with any other task. As before, we split the subjects randomly into train and test sets of size 80 and 20 respectively. For each subject in the train set, we find the features corresponding to the top leverage scores. This gives us 80 sets of features. We find the set of statistically significant features for the entire training set using a hyper-geometric p-value test (threshold $10^{-50}$) to identify recurring features across all subjects. These statistically significant features are used to restrict the feature space in the test set. The accuracy, in terms of correctly identified tasks in the test set, averaged across 1000 trials was $91.93$ ($\pm 1.34\%$). In contrast, the same experiment using the entire connectome yielded accuracy of $84.95$ ($\pm 2.2\%$). 

\begin{figure*}
    \centering
    \includegraphics[scale=0.3]{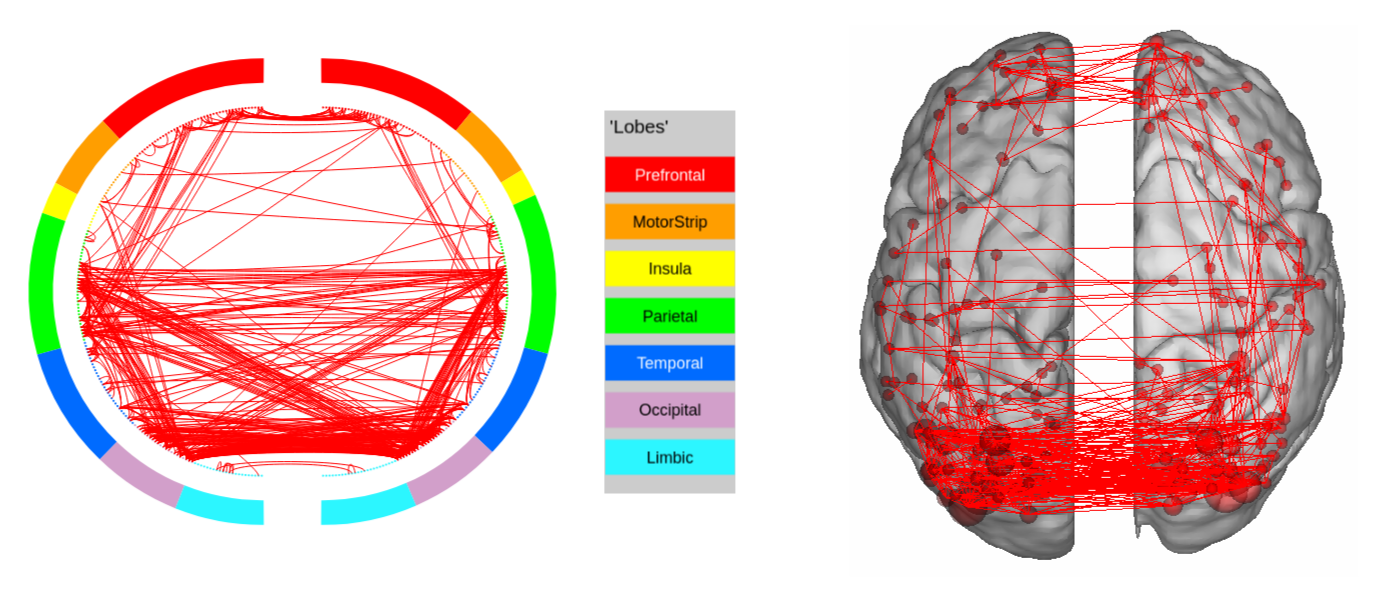}
    \caption{{\bf Task identifiability for a single subject.} These results show that the edges that distinguish tasks in the Human Connectome Project are strongly localized to the occipital lobe. }
    \label{fig:task_identifiability}
\end{figure*}

We also observe through this experiment that the regions expressing these task-identifying signatures are strongly localized to the cortex. In Figure \ref{fig:task_identifiability}, we show the high confidence edges. For clarity of visualization, we only show edges where at least one of the end nodes has a sufficiently high degree (> 30). This figure illustrates our hypothesis clearly. A significant fraction of the high confidence edges are connected to the Occipital region, with a strong connectivity to the Parietal region. The Occipital region contains the different layers of the visual cortex. These layers processes various visual stimuli that a subject is exposed to, during the course of the different tasks. Similarly, the Parietal lobe is responsible for integrating sensory information for the visual cortex. Our results suggest that the signature captures the difference in processing different stimuli, which is integral to the description of any task. Finally, we note that the regions are dependent on the task set: for instance, the Occipital lobe is not significantly implicated in a set of tasks with no visual stimulus.

\section*{Discussion}
In this paper, we use a novel matrix sampling method to answer critical questions related to individual and task-specific signatures in the human brain connectome. We show that a very small number of features in a functional connectome codes for the identity of an individual. Furthermore, we show that these features (edges in the connectome) are robust, statistically significant, and invariant across individuals. The regions corresponding to these features are consistent with existing literature -- both for resting-state and task, strongly supporting the physiological basis for our method. We also show that the features that code for specific tasks are localized strongly to the Occipital Lobe.

The core contribution of our method lies in the fact that it can predict relevant features, without requiring multiple scans. In doing so, it generalizes the application of brain signatures beyond the HCP dataset. Specifically, it can be used in any study, where the goal is to find individual differences in a cohort. Furthermore, since the method is built on sound theoretical foundation, its computational and statistical underpinnings are well-characterized.

The success of leverage scores in picking relevant features for brain signatures suggests that it can also be used in conventional fMRI studies, where the aim is to find differences between cases and controls. The key distinction between the two setups is that the former requires unsupervised feature selection, whereas the latter requires supervised feature selection. We note that this necessitates new sampling techniques, along with associated theoretical guarantees.
\section*{Methods and Materials}
\subsection*{Dataset}

The images used in this study are collected as part of the Healthy Young Adult study in the WU-Minn Project \cite{VanEssen13} by the Human Connectome Project Consortium. The data includes 3T structural and functional MRI for 1113 adults, 7T resting and task magnetoencephalography (MEG) from 184 subjects, and 3T and 7T diffusion data. Full details of the data can be found in vanEssen et al.~\cite{VanEssen12}. 

The resting state functional MRI is acquired in two sessions, on two separate days. The resting state data in each session lasted approximately 30 minutes (15 minutes for L-R and R-L phase encoding). The spatial resolution was $2 \times 2 \times 2\ mm^3$ and temporal resolution (TR) of 720 ms. In this study, we use the L-R encoding from the 100 ``unrelated'' subjects, which is part of the S900 release. For a detailed description of the acquisition protocol see Smith et al.~\cite{Smith13}. 

In each session, the resting state session was followed by tasks. On the first day, the tasks included working memory, gambling and motor; the second day included language, social cognition, relational processing and emotional processing. The task functional MRI data also have the L-R encoding and the R-L encoding. However, the duration of each session varied considerably, ranging from 176 frames in emotion processing to 401 frames in working memory, with TR of 5520 ms. For a detailed description of the protocol see Barch et al.~\cite{Barch13}.

\subsubsection*{Preprocessing}
\label{subsection_preprocessing}

The preprocessing steps follow the minimum pre-processing pipeline prescribed by Glasser et al.~\cite{Glasser13} and Smith et al.~\cite{Smith13}. The HCP pre-processing steps include spatial pre-processing and temporal pre-processing. In resting state functional MRIs, we also perform global signal regression. 

The goal of spatial pre-processing is to remove artifacts, while leaving useful information intact. It includes correction for spatial distortions due to gradient non-linearity (of magnetic field), followed by correction for head motion of subject during data acquisition, and registration to the structural MRI image. Since brains of different subjects are of different sizes, the image is transformed to 2mm MNI co-ordinate system. This is followed by normalizing the 4D image to global mean and masking out non-brain voxels. Finally, the cortical surface vertices and the sub-cortical volumetric vertices are combined into the standard grayordinate space in a CIFTI dense time-series format. The surface cortical voxels are aligned using non-linear surface registration and non-linear volume registration for volume structures, thereby achieving spatial correspondence between images of different subjects. The rationale behind the grayordinate space is explained in Glasser et al.~\cite{Glasser13}. In this study, we only use time series from cortical vertices.

Temporal pre-processing is particularly important for resting state analyses, as they rely on correlation between voxel time series. These can get corrupted by temporal artifacts that span across several voxels (in contrast, task analyses have pre-determined timed events, which make them robust to temporal artifactual influences). The HCP temporal pre-processing pipeline includes a minimal high pass filter (cutoff of 2000s) and a slow roll off below that point, which achieves a linear de-trending of data. For task functional MRI, the cutoff was 200s.

In the case of resting state functional MRI, we follow this by global signal regression, a procedure that regresses out the mean time series out of every voxel. Since resting state analyses have shown that fluctuations at low frequencies are due to haemodynamic responses to neural activation, we applied a bandpass filter (0.008Hz to 0.1Hz).

\subsubsection*{Brain Atlas}
\label{subsection_atlas}

The description of a functional connectome as a network requires the identification of a set of nodes and a set of edges between these nodes. A brain atlas defines parcellation schemes that help achieve the first requirement. Ideally, parcels are non-overlapping, locally functionally distinct areas. The variation within a parcel should be dominated by the variation between parcels. While parcellation is an abstraction, it allows for analyses of a structurally and functionally relevant  region $\times$ region network, rather than a raw {\it voxel $\times$ voxel} network.

In our work, the cortical structures are parcellated in accordance with the atlas of Glasser et al.~\cite{Glasser16}. This atlas consists of 360 regions (180 in each hemisphere), bound by sharp edges on the basis of anatomy, function, and/or topology. Each of the 180 regions are also classified into 22 larger regions, which allows for analysis at a coarser level, if needed.  We find that other parcellation schemes either have too few regions, or are annotated on the basis of fewer samples. In contrast, the boundaries in Glasser et al.~\cite{Glasser16} are drawn using resting state and task fMRI images from two groups of 210 subjects, all of which is part of the previously described Human Connectome Project.

\subsection*{Methodology}

\subsubsection*{Experimental Setup}
\label{subsection_setup}

The time series data (i.e., a 2D matrix of gray-ordinate {\it voxel $\times$ time-points}) is z-score normalized. Following this, voxels belonging to the same parcel 
are averaged to yield a {\it region $\times$ time-points} matrix. The Pearson Correlation of all pairs of time series is then computed, resulting in a {\it region $\times$ region} correlation matrix. This process is repeated for both sessions of all subjects. Then, the upper triangular matrix for the first session for all subjects is vectorized and stacked next to each other into a {\it subjects $\times$ feature} matrix, where feature is an entry in the correlation matrix. A similar matrix is constructed for the second session as well. 

We first discuss the concept of row and column sampling as a general means of feature selection and, in particular, leverage score sampling. We then discuss the use of leverage scores to select features for brain fingerprinting. 

\subsubsection*{Row/column sampling}
\label{subsection_sampling}

Given matrix $A$, an individual entry $a_{i,j}$ corresponds to the weight of the $i$th edge of subject $j$. The problem of identifying the most discriminating subset of features then translates to selecting non-zeros from the correlation matrix (row-column pairs) that are most discriminative in terms of individual signatures. In contrast to conventional dimensionality reduction techniques such as Singular Value Decomposition (SVD),
the ability to choose rows or columns from a data matrix directly translates to \textit{feature selection}, especially when features have physical meaning. Retaining such features in the matrix sketch can make the underlying physical phenomenon explainable, while also de-noising the data by eliminating non-discriminating features and noise. Furthermore, in contrast to methods that use within-subject and between-subject metrics to reduce dimensions as in Byrge et al.~\cite{Byrge19}, we do not require data from both sessions to identify important edges.

Given a matrix $A \in \mathbb{R}^{m \times n}$, one can use the following randomized meta algorithm to create a sketch matrix $\tilde{A}$ that retains $s$ ($< m$) rows. 

\begin{center}
\begin{algorithmic}
\Function {row\_sample} {A,s}
\State {Let $\tilde{A}$ be an empty matrix}
\For {$t = 1$ to $s$}
\State {Randomly sample a row according to the distribution $P$}
\State {Let $A_{i_t,\star}$ be the sampled row, with corresponding probability $p_i$}
\State {Set $\tilde{A}_{t,\star} = \frac{1}{\sqrt{s p_i} }A_{i_t,\star}$}
\EndFor\\
\Return $\tilde{A}$
\EndFunction
\end{algorithmic}
\end{center}

The algorithm samples $s$ rows of $A$ in independent, identically distributed trials according to $P$s. The re-scaling of $\tilde{A}$ ensures that $\tilde{A}^T \tilde{A}$ is unbiased, i.e., $\mathbb{E}[(\tilde{A}^TA)_{i,j}] = (A^T A)_{i,j}, \forall i \in \{ 1 \dots m\}, j \in \{ 1 \dots n\}$ \cite{Drineas06}.

The only unspecified detail is the choice of distribution $P$. A simple choice is to sample rows uniformly, however, this yields poor results. An intuitively better choice for the distribution relies on the matrix $A$ itself -- assigning higher weights to more important elements. A simple non-uniform distribution is based on $l_2$ sampling, which can be defined as:
\begin{equation}
p_i = \frac{|| A_{i,\star} ||_2^2}{\sum_i || A_{i,\star} ||_2^2} = \frac{|| A_{i,\star} ||_2^2}{|| A_{i,\star} ||_F}.
\end{equation}
Using norm-squared sampling, Drineas et al.~\cite{Drineas06}  prove that: 
\begin{equation}
\mathbb{E}[|| A^T A - \tilde{A}^T\tilde{A}||_F] \leq \frac{1}{\sqrt{s}}||A||_F^2.
\label{two_norm_ineq}
\end{equation}
The bounds in Equation~\ref{two_norm_ineq} imply that  $\tilde{A}$ can be used as a proxy for $A$. However, this approximation introduces an additive error, which depends on $||A||_F$. To achieve better bounds, one can make use of the knowledge of column space of $A$. The associated sampling technique is called leverage score sampling~\cite{Drineas16}.

\paragraph{Leverage Score Sampling.}
Let $A \in \mathbb{R}^{m \times n}$, with $m \gg n$. Let $U \in \mathbb{R}^{m \times n}$ be any orthonormal matrix spanning the column space of $A$. Then, $U^T U = I$ and $U U^T = P_A$, which is an $m$-dimensional projection matrix onto the span of $A$. Then, the probabilities $p_i$ are defined as:
\begin{equation}
p_i = \frac{|| U_{i,\star} ||_2^2}{\sum_i || U_{i,\star} ||_2^2}= \frac{1}{n}(P_A)_{i,i} \qquad \forall i \in \{1 \dots m\}.
\label{leverage_probability_eqn}
\end{equation}
The values of $p_i$s in Equation \ref{leverage_probability_eqn} are known as \textit{statistical leverage scores}. 

In our matrix of vectorized edge values, each row contains values expressing functional connectivity markers between the same pair of physical regions on the brain cortex. Hence, leverage scores are indicative of relative importance of edges in discriminating samples. While the randomized approach discussed here helps in understanding the process, we find that a deterministic approach performs well in practice. We call this sketching process \textit{Principal Features Subspace Method}.

\subsubsection*{Principal Features Subspace Method}
\label{subsection_principal_features_subspace}

As before, let $A$ be the data matrix of connectomes, and $U$ be the orthonormal matrix that spans the column space of $A$. Additionally, let $t$ be the number of features that need to be retained. An example of such a matrix $U$ is constructed using left singular vectors from a Singular Value Decomposition (SVD) of $A$. We can then compute the leverage($l$) scores of $A$ as:
\begin{equation}
l_i = || U_{i,\star} ||_2^2, \qquad \forall i \in \{1 \dots m\}.
\end{equation}

We sort the leverage scores and retain the features corresponding to the top $t$ leverage scores. We call this subspace the \textit{principal features subspace}. In contrast to prior randomized approaches, we select features in a deterministic manner; Cohen et al.~\cite{Cohen15} provide theoretical bounds for this selection process.

Starting from the matrix of vectorized edge values $A$, we compute the left singular vectors using SVD. The ordering of edges according to their leverage scores, if robust across different groups, is indicative of a set of features  that can accurately fingerprint an individual's functional connectome. In this case, for a given parcellation scheme, and for a given measure of region-region coherence, we need to apply SVD just once to determine the relevant edges. We present results from our scheme applied to 100 MRI samples to demonstrate powerful new results on the compactness and robustness of the sub-connectome coding individual fingerprints.

\subsection*{Related Methods}

There is significant recent work in characterizing population-level differences from neuroimaging datasets. Given the focus of our work on individual-level differences, we restrict ourselves to significant results in this space, and use them to motivate our new results.

Finn et al.~\cite{Finn15} divide the identification task into two steps: whole brain and network specific. In the first step, they perform identification on the basis of the region-wise correlation matrix of the whole brain, and report a success rate of approximately 93\%. Following this, they divide the brain into eight functional networks: medial frontal, fronto-parietal, default mode, subcortical-cerebellum, motor, visual I, visual II, and visual association, on the basis of an atlas constructed from the Yale Dataset (268 nodes covering the whole brain). Restricting analysis to each of these regions, they perform the identification task. From this analysis, they find that the networks medial frontal and fronto parietal carry most discriminating features, yielding accuracies as high as 98\%. Similar results have also been reported by Miranda-Dominguez et al.\cite{Miranda14} and Mueller et al.~\cite{Mueller13}. However, these results require significant prior knowledge on composite of parcels known to have functional/structural coherence. 

Our effort takes a similar two-level approach, viewing identification at the whole brain and network levels. Instead of using prior knowledge, we show that the \textit{Principal Features Subspace} yields a comparable identification training accuracy of approximately 96\%. Using the features obtained in the training set on the test set, we find that our test accuracy is around 93\%, which is significantly higher than that from Finn et al. \cite{Finn15}. In contrast to Finn et al. we restrict the analysis in the second step to the regions that are computationally identified from our first step. The regions obtained from our first step are in fact consistent with those used by Finn et al. in the second step, in that we find regions in the parietal and frontal cortex being over-represented in our top nodes (9 out of 12 belong to fronto-parietal regions). However, the network identified by our method is significantly smaller than the region-specific network of Finn et al. It is shown to be robust, statistically significant, and invariant on test subjects. Restricting ourselves to 24 parcels (which may belong to different networks), we report an accuracy of approximately 96\%. 

There have been recent attempts aimed at identifying discriminating edges, using the so-called {\it differential power} introduced by Finn et al. and improved greatly by Byrge et al. \cite{Byrge19}. In comparison to Byrge et al., our approach has two distinct advantages. First, our \textit{principal features subspace} method uses the concept of leverage scores which has strong theoretical guarantees \cite{Drineas06, Drineas16, Cohen15, Papailiopoulos14}; these guarantees are not provided by {\it differential power}. Second, our approach does not require any labels in order to find discriminating edges. This significantly increases the applicability of our method, since most studies do not have multiple scans of a subject. As acquisition protocols may vary from study to study, it isn't possible to use the features from HCP in other studies, where it may be necessary to find subject-level unique patterns. Our setup simply requires two sets of images, with the guarantee that each subject has one scan in each set. We use only one set to find discriminative edges and the second set to confirm the efficacy of our approach. In this context, the definition of differential power is not applicable.

Column/row sampling using randomized techniques have significant benefits, in addition to de-noising and avoiding overfitting.  Other commonly used de-noising techniques transform the data matrix into a combination of linearly independent (orthogonal) components, using methods such as Principal Component Analysis (PCA) or Singular Value Decomposition (SVD). De-noising can be achieved by discarding the components corresponding to noise, and working with a low rank approximation of the matrix, as explored in context of brain fingerprinting by Amico et al.~\cite{Amico17}.

Using the ``unrelated'' subset of HCP subjects, Amico et al. \cite{Amico17} report that using the original data matrix  (simply computing correlations with known samples and selecting the most correlated subject for identification) as a fingerprint yielded an accuracy of around 87\%. The process of de-noising using low rank approximations yields better results -- Amico et al.\cite{Amico17} report an accuracy of 91\% for brain fingerprinting resting state fMRIs. 

As discussed earlier, the advantage of the \textit{principal features subspace} method is that it provides a small set of regions, thereby de-noising data while also making the connectivity matrix much smaller. It also identifies the signature as a set of biologically relevant features.

\bibliography{bibliography.bib}
\bibliographystyle{plainnat}  

\end{document}